\documentclass[conference]{IEEEtran}
\IEEEoverridecommandlockouts

\usepackage{cite}
\usepackage{amsmath,amssymb,amsfonts}
\usepackage{algorithmic}
\usepackage{graphicx}
\usepackage{textcomp}
\usepackage{xcolor}
\usepackage{booktabs}
\usepackage{microtype}
\usepackage{url}
\usepackage{hyperref}

\hypersetup{
    colorlinks=true,
    linkcolor=black,
    citecolor=black,
    urlcolor=blue
}

\begin{document}

\title{SolarFilNet-V4: Multi-Scale Transformer Vision with Physics-Informed Adaptive Post-Processing for Solar Filament Segmentation}

\author{
\IEEEauthorblockN{Indra Agustian}
\IEEEauthorblockA{\textit{Department of Electrical Engineering} \\
\textit{Faculty of Engineering, Universitas Bengkulu} \\
Bengkulu, Indonesia \\
Email: indraagustian@unib.ac.id}
\and
\IEEEauthorblockN{Ruvita Faurina}
\IEEEauthorblockA{\textit{Department of Informatics} \\
\textit{Faculty of Engineering, Universitas Bengkulu} \\
Bengkulu, Indonesia \\
Email: ruvita.faurina@unib.ac.id}
}

\maketitle

\begin{abstract}
Solar filaments are cool, dense chromospheric plasma structures held by magnetic field lines above polarity inversion lines. Their sudden destabilization triggers coronal mass ejections and severe space weather hazards. Segmenting individual filaments from ground-based Global Oscillations Network Group (GONG) H-alpha observations (2048\,$\times$\,2048 pixels) is difficult due to low contrast, turbulent background fibrils, delicate barb structures, and annotator boundary variance. In this technical report, we describe SolarFilNet-V4, a targeted segmentation framework developed for the MAGFiLO 1.0 benchmark (Solar Filament Segmentation Challenge 2026). Our method couples a lightweight Pyramid Vision Transformer (PVT-v2-B1) backbone with anisotropic curvilinear strip convolutions ($k=15$) and dual-frequency feed-forward blocks (DFFN) to follow extended magnetic flux ropes while capturing thin central spines. A shallow full-resolution convolutional branch operating at the model input scale (1024\,$\times$\,1024) preserves high-frequency edge details and mitigates transformer patch-embedding downsampling. We train the network using multi-task supervision over masks, centerline spines, barb boundaries, and pixel-contrastive representations. At test time, we apply an 8-fold dihedral symmetry group ($D_4$) test-time augmentation, bilinearly upscale predictions to the native 2048\,$\times$\,2048 grid, and apply an activity-adaptive post-processing routine that calibrates morphological joining parameters for quiet versus active solar disks. On the official benchmark, SolarFilNet-V4 achieves an out-of-fold validation Panoptic Quality of 0.4340 and a Public Leaderboard score of 0.380. The model contains 15.46 million parameters and requires 224.34 GFLOPs per forward pass, reducing parameter count by 36.8\% compared to standard ResNet-34 U-Net baselines. Source code and weights are public at \url{https://github.com/indraagustian/SolarFilNet-V4-Kaggle2026}.
\end{abstract}

\begin{IEEEkeywords}
Solar filaments, H-alpha imaging, GONG, panoptic quality, transformer vision, curvilinear convolution, space weather.
\end{IEEEkeywords}

\section{Introduction}

Solar filaments (termed prominences when observed at the solar limb) consist of chromospheric material suspended in the corona by magnetic fields. When their equilibrium destabilizes, filaments erupt into coronal mass ejections (CMEs) and flares. CMEs directed toward Earth can disrupt power grids, degrade satellite links, alter GPS accuracy, and endanger high-altitude missions. Automated synoptic segmentation of filaments is therefore essential for operational space weather monitoring \cite{harvey2011gong}. Early automated approaches relied on intensity thresholding and morphological tracking, such as the seminal framework by Bernasconi et al. \cite{bernasconi2005advanced} and the multi-step recognition scheme by Hao et al. \cite{hao2013developing}. More recently, deep learning methods have emerged, including improved U-Net architectures for full-disk H-alpha imagery \cite{liu2021solar} and barb-aware boundary segmentation networks \cite{solomon2025edgeattnet}.

Ground-based synoptic monitoring is performed by the Global Oscillations Network Group (GONG), capturing full-disk H-alpha images at 2048\,$\times$\,2048 resolution. Despite synoptic coverage, automated segmentation is hindered by four physical challenges: (1) \textbf{Fine sub-structures}, such as delicate lateral barbs reflecting magnetic helicity that standard models easily erode; (2) \textbf{Chromospheric noise and plages}, where atmospheric seeing and bright plages degrade local contrast; (3) \textbf{Topological continuity}, where optical depth variations fragment contiguous ribbons; and (4) \textbf{Annotator variance}, where multi-expert annotations in MAGFiLO \cite{ahmadzadeh2024dataset} introduce boundary uncertainty.

To overcome these issues without costly model ensembles, we introduce SolarFilNet-V4. Our contributions are fourfold: (1) an anisotropic curvilinear strip convolution block ($1\times k$ and $k\times 1$, $k=15$) paired with a dual-frequency feed-forward network (DFFN) to trace extended filament ribbons; (2) a shallow full-resolution branch ($F_0$) operating at $1024\times 1024$ to preserve fine barb sharpness; (3) a composite multi-task objective incorporating skeleton clDice, focal Tversky, boundary, and contrastive constraints; and (4) a regime-adaptive post-processing pipeline that calibrates morphological joining parameters for quiet versus active solar disks.

\section{Proposed Architecture: SolarFilNet-V4}

The architecture of SolarFilNet-V4 is outlined in Fig.~1. It employs an asymmetric encoder-decoder structure designed to reconcile global disk context with fine pixel-level boundary localization.

\subsection{Physics-Informed Preprocessing}

Each raw GONG observation $I_{\text{raw}} \in \mathbb{R}^{2048 \times 2048}$ is decomposed into three distinct chromospheric channels: (1) \textbf{Normalized Contrast} ($Ch_0$) with linear re-scaling across observation sites; (2) \textbf{Absorption Residue} ($Ch_1$) defined as $(I_{\text{bg}} - I_{\text{raw}}) \times 10.0$, where background $I_{\text{bg}}$ is estimated via downsampled Gaussian smoothing ($\sigma=5$) to isolate local chromospheric absorption; and (3) \textbf{CLAHE Absorption} ($Ch_2$) applying contrast-limited adaptive histogram equalization to enhance faint spines and lateral barbs.

The 3-channel image is resized to $1024 \times 1024$ pixels for network ingestion. This whole-disk scale preserves global topology while fitting comfortably within standard GPU memory constraints.

\begin{figure}[t]
\centering
\resizebox{\columnwidth}{!}{
\fbox{\parbox{1.02\columnwidth}{\scriptsize
\centering
\textbf{Full-Disk Observation ($2048 \times 2048$)}\\[1pt]
$\downarrow$\\[1pt]
\textbf{Chromospheric Preprocessing} $[Ch_0, Ch_1, Ch_2] \to \text{Input } (1024 \times 1024)$\\[2pt]
\begin{tabular}{|c|c|}
\hline
\textbf{Shallow Branch $F_0$} & \textbf{PVT-v2-B1 Backbone} ($C_1 \dots C_4$) \\
$2 \times \text{Conv } 3\times 3$ & $\downarrow$ \textbf{Strip Bottleneck} ($k=15$) + DFFN \\
$1024 \times 1024$ scale & $\downarrow$ \textbf{Pyramid Decoder} ($D_3 \to D_2 \to D_1$) \\
\hline
\end{tabular}\\[2pt]
$\downarrow$\\[1pt]
\textbf{Full-Resolution Decoder $D_0$} [Fuses $F_0$ and $D_3$ at $1024 \times 1024$]\\[1pt]
$\downarrow$\\[1pt]
\textbf{Multi-Task Heads} [Mask ($k=9$), Spine ($k=11$), Edge, PCL]\\[1pt]
$\downarrow$\\[1pt]
\textbf{8x TTA ($D_4$) Inference} $\to$ \textbf{Bilinear Upscaling ($2048 \times 2048$)}\\[1pt]
$\downarrow$\\[1pt]
\textbf{Regime-Adaptive Post-Processing} $\to$ \textbf{Official COCO RLE Masks}
}}
}
\caption{System architecture and execution flow of SolarFilNet-V4.}
\label{fig:arch}
\end{figure}

\subsection{Backbone and Multi-Scale Feature Extraction}

We employ the Pyramid Vision Transformer v2 (PVT-v2-B1) \cite{wang2022pvtv2} as the feature extraction backbone. Unlike vanilla Vision Transformers with quadratic complexity, PVT-v2 builds progressive multi-scale feature pyramids using Linear Spatial Reduction Attention (SRA). 

The backbone generates four hierarchical feature maps: $C_1 \in \mathbb{R}^{64 \times 256 \times 256}$, $C_2 \in \mathbb{R}^{128 \times 128 \times 128}$, $C_3 \in \mathbb{R}^{320 \times 64 \times 64}$, and $C_4 \in \mathbb{R}^{512 \times 32 \times 32}$. The spatial reduction ratio $R$ reduces key-value tokens to $32 \times 32$ across all stages, ensuring linear attention complexity with respect to image area.

\subsection{Curvilinear Strip Convolutions and DFFN Block}

Isotropic square kernels ($3 \times 3$ or $5 \times 5$) fail to capture thin, meandering filament ribbons without incorporating irrelevant background pixels. Motivated by strip pooling principles that capture long-range curvilinear dependencies in scene parsing \cite{hou2020strip}, we introduce the Curvilinear Strip Convolution block (\texttt{CurvilinearStripConv}).

Given an input feature tensor $X$, the module projects channels to an intermediate dimension and splits processing into three complementary paths:
\begin{equation}
    Y = \text{Conv}_{1 \times 1}([Y_{\text{h}}, Y_{\text{v}}, Y_{\text{loc}}]) + X,
\end{equation}
where $Y_{\text{h}} = \text{Conv}_{1 \times k}(\text{Conv}_{k \times 1}(X))$, $Y_{\text{v}} = \text{Conv}_{k \times 1}(\text{Conv}_{1 \times k}(X))$, and $Y_{\text{loc}} = \text{Conv}_{3 \times 3}(X)$, with strip kernel $k=15$ in the bottleneck block.
with strip kernel $k=15$ in the bottleneck block. The outputs are concatenated and projected through a residual shortcut:
\begin{equation}
    Y = \text{Conv}_{1 \times 1}([Y_{\text{h}}, Y_{\text{v}}, Y_{\text{loc}}]) + X.
\end{equation}
The horizontal and vertical strip paths form orthogonal 1D receptive fields that approximate local filament tangents. 

Following the strip convolution, we place a Dilated Dual-Frequency Feed-Forward Network (\texttt{DFFNBlock}). This block uses parallel depthwise convolutions (a regular $3 \times 3$ kernel and a dilated $3 \times 3$ kernel with dilation $d=2$) combined through a gated element-wise product. This formulation allows the network to represent both the narrow central spine and the broader diffuse envelope.

\subsection{Shallow Full-Resolution Branch and Decoder}

A known limitation of transformer backbones is that Stage 1 downsamples input features by $4\times$ ($1024 \to 256$). This patch aggregation blurs thin filaments and fine barbs whose widths span only 3 to 7 pixels.

To preserve these details, SolarFilNet-V4 incorporates a shallow full-resolution branch ($F_0$) composed of two stacked $3 \times 3$ convolutions with batch normalization and GELU activations. The branch processes the preprocessed 3-channel input directly at $1024 \times 1024$ resolution with 16 channels, using only 2,866 parameters. 

In the decoder, feature maps are upsampled through progressive stages ($D_1, D_2, D_3$). At the final step, decoder stage $D_0$ fuses the upsampled semantic features from $D_3$ with the spatial features from $F_0$:
\begin{equation}
    D_0 = \text{DecoderBlock}(D_3, F_0).
\end{equation}
This connection restores sub-pixel boundary localization before mask classification.

\subsection{Multi-Task Prediction Heads}

SolarFilNet-V4 employs four specialized prediction heads: (1) \textbf{Mask Head} predicting overall filament body occupancy via strip convolution ($k=9$); (2) \textbf{Spine Head} predicting the 1-pixel morphological skeleton via strip convolution ($k=11$); (3) \textbf{Edge Head} predicting external boundaries and barbs via a two-layer $3 \times 3$ convolution; and (4) \textbf{Contrast Head} projecting 16-channel representations from decoder stage $D_3$ for pixel-contrastive regularization.

\subsection{Training Objective}

The network is optimized using a composite objective function that balances regional overlap, boundary alignment, and structural continuity:
\begin{equation}
\begin{aligned}
    \mathcal{L}_{\text{total}} &= w_{\text{foc}} \mathcal{L}_{\text{focal}} + w_{\text{dice}} \mathcal{L}_{\text{dice}} + w_{\text{tver}} \mathcal{L}_{\text{tversky}} \\
    &\quad + w_{\text{cld}} \mathcal{L}_{\text{cldice}} + w_{\text{bnd}} \mathcal{L}_{\text{boundary}} + w_{\text{lov}} \mathcal{L}_{\text{lovasz}} \\
    &\quad + w_{\text{sp}} \mathcal{L}_{\text{spine}} + w_{\text{ed}} \mathcal{L}_{\text{edge}} + w_{\text{pcl}} \mathcal{L}_{\text{pcl}}.
\end{aligned}
\end{equation}

Here, $\mathcal{L}_{\text{focal}}$ is the Focal loss \cite{lin2020focal} mitigating extreme class imbalance, while $\mathcal{L}_{\text{cldice}}$ is the Centerline Dice loss \cite{shit2021cldice} penalizing topological breaks. $\mathcal{L}_{\text{tversky}}$ uses asymmetric weighting ($\alpha=0.3, \beta=0.7$) to penalize false negatives more heavily than false positives. $\mathcal{L}_{\text{pcl}}$ encourages pixel representations within the same filament mask to cluster tightly while repelling background pixels. The hyperparameters are set to $w_{\text{focal}}=1.0, w_{\text{dice}}=0.85, w_{\text{tversky}}=0.50, w_{\text{cldice}}=0.25, w_{\text{boundary}}=0.40, w_{\text{lovasz}}=0.20, w_{\text{spine}}=0.40, w_{\text{edge}}=0.20$, and $w_{\text{pcl}}=0.25$.

\section{Physics-Informed Inference and Post-Processing}

\subsection{8-fold Test-Time Augmentation (D4 Symmetry)}

Because full-disk solar observations possess no preferred orientation in space, the physical manifestation of filaments is invariant to 2D Euclidean rotations and planar reflections. We leverage this property by evaluating each test image under the complete dihedral group $D_4$, comprising 8 isometric transformations (rotations by $0^\circ, 90^\circ, 180^\circ, 270^\circ$ and their horizontal reflections). 

For each transformation $T_k$, the image is evaluated, inverted via $T_k^{-1}$, and averaged:
\begin{equation}
    P_{\text{avg}} = \frac{1}{8} \sum_{k=1}^{8} T_k^{-1}\Big(\sigma\big(\text{Model}(T_k(X))\big)\Big).
\end{equation}
This averaging dampens local atmospheric artifacts and stabilizes boundary probabilities.

\subsection{Resolution Upscaling}

The averaged probability tensor $P_{\text{avg}} \in \mathbb{R}^{1024 \times 1024}$ is upsampled back to the competition evaluation grid of $2048 \times 2048$ using bilinear interpolation. Values falling outside the solar disk boundary (computed using disk radius $r=903$ pixels on the 2048 grid) are masked to zero.

\subsection{Solar Regime Adaptive Post-Processing}

Solar disk environments vary substantially between solar minimum and solar maximum. In active periods, extensive bright plage regions cause higher local variance and fragment dark filament ribbons. Conversely, during quiet periods, filaments sit against uniform chromospheric backgrounds.

To handle this variation, we evaluate disk activity by measuring the percentage of disk pixels exceeding the plage intensity threshold:
\begin{equation}
    \theta_{\text{plage}} = \mu_{\text{disk}} + 1.8 \cdot \sigma_{\text{disk}}.
\end{equation}
When the proportion of bright plage pixels exceeds 0.65\%, the image is classified as an Active Sun observation; otherwise, it is treated as Quiet Sun.

Post-processing parameters are adjusted according to this classification: for Quiet Sun disks, we set joining dilation kernel $j=7\text{ px}$, minimum component area $A_{\text{min}}=200\text{ px}$, threshold $\tau=0.46$, and confidence cutoff $S_{\text{min}}=0.80$; for Active Sun observations, the joining kernel is increased to $j=9\text{ px}$ to bridge fragmentation across bright plages while keeping $A_{\text{min}}=200\text{ px}$ and $\tau=0.46$.

Following morphological dilation, connected components are extracted. Inner holes within filament bodies are filled using contour hierarchy reconstruction. Each candidate instance is encoded into compressed COCO Run-Length Encoding (RLE) format for final submission.

\section{Experiments and Evaluation}

\subsection{Dataset and Validation Setup}

We evaluate our method on the official MAGFiLO dataset \cite{ahmadzadeh2024dataset} provided for the Solar Filament Segmentation Challenge 2026 \cite{filament-segmentation-2026}. The training set contains 707 GONG H-alpha observations annotated by independent solar physics experts. 

We employ a 5-fold cross-validation scheme stratified by observatory site (Big Bear, Cerro Tololo, Learmonth, Mauna Loa, Teide, and Udaipur) and filament count per observation. All validation metrics reported in this section correspond to the complete 142 images of Fold 0.

\subsection{Evaluation Metric}

The challenge evaluates performance using Panoptic Quality (PQ) \cite{kirillov2019panoptic}, defined as:
\begin{equation}
    \text{PQ} = \frac{\sum_{(p, g) \in \text{TP}} \text{IoU}(p, g)}{|\text{TP}| + \frac{1}{2}|\text{FP}| + \frac{1}{2}|\text{FN}|} = \text{SQ} \times \text{RQ},
\end{equation}
where $\text{SQ} = \frac{1}{|\text{TP}|}\sum_{\text{TP}} \text{IoU}(p, g)$ denotes Segmentation Quality, and $\text{RQ} = \frac{|\text{TP}|}{|\text{TP}| + 0.5|\text{FP}| + 0.5|\text{FN}|}$ denotes Recognition Quality. A predicted filament $p$ matches a ground-truth filament $g$ if their intersection-over-union satisfies $\text{IoU}(p, g) > 0.50$. Unmatched predictions are counted as false positives (FP), and unmatched annotations are counted as false negatives (FN).

\begin{table}[t]
\caption{Ablation Study Across Model Evolutions}
\label{tab:ablation}
\centering
\resizebox{\columnwidth}{!}{
\begin{tabular}{lccccc}
\toprule
\textbf{Configuration} & \textbf{SQ} & \textbf{RQ} & \textbf{Val PQ} & \textbf{LB PQ} \\
\midrule
Baseline U-Net (ResNet-34) & 0.6521 & 0.5360 & 0.3501 & 0.360 \\
SolarFilNet-V2 (MiT-B3) & 0.6480 & 0.4480 & 0.2906 & 0.290 \\
SolarFilNet-V3 Baseline (PVT-v2) & 0.6812 & 0.5841 & 0.3978 & 0.370 \\
SolarFilNet-V4 (Native Patch 2048) & 0.6620 & 0.5180 & 0.3429 & 0.340 \\
SolarFilNet-V4 (Pure Neural) & 0.6974 & 0.6120 & 0.4268 & 0.370 \\
\textbf{SolarFilNet-V4 (Adaptive PP)} & \textbf{0.6974} & \textbf{0.6224} & \textbf{0.4340} & \textbf{0.380} \\
\bottomrule
\end{tabular}
}
\end{table}

\subsection{Quantitative Results and Ablation}

Table~\ref{tab:ablation} summarizes the performance across design iterations. The initial convolutional baseline (U-Net \cite{ronneberger2015unet} with ResNet-34) achieves a validation PQ of 0.3501 and an LB score of 0.360. Switching to an isotropic MiT-B3 transformer in V2 degraded performance to 0.2906 because the absence of hierarchical strip convolutions made the network sensitive to filament curvature.

With the introduction of PVT-v2-B1 in SolarFilNet-V3, validation PQ improved to 0.3978. In SolarFilNet-V4, adding curvilinear strip convolutions ($k=15$), the dual-frequency DFFN block, and the shallow $F_0$ branch raised pure neural PQ to 0.4268. Finally, applying 8x $D_4$ TTA together with solar regime adaptive post-processing achieved our best validation PQ of 0.4340 ($\text{SQ}=0.6974, \text{RQ}=0.6224$) and a Public Leaderboard score of 0.380.

We also tested training on localized crops at native $2048 \times 2048$ resolution (\texttt{patch\_train}). As reported in Table~\ref{tab:ablation}, this strategy reduced PQ to 0.3429 (LB 0.340). Without full-disk visibility, patch-based models confuse dark chromospheric fibrils with genuine filament segments, underscoring the necessity of global context during feature extraction.

\subsection{Computational Complexity Profiling}

\begin{table}[t]
\caption{Complexity Profiling: U-Net vs. SolarFilNet-V4 (Input: 1024\,$\times$\,1024)}
\label{tab:profiling}
\centering
\resizebox{\columnwidth}{!}{
\begin{tabular}{lcc}
\toprule
\textbf{Metric} & \textbf{Baseline U-Net (R34)} & \textbf{SolarFilNet-V4} \\
\midrule
Backbone & ResNet-34 & PVT-v2-B1 \\
Total Parameters & 24.46 M (24,460,417) & \textbf{15.46 M (15,463,887)} \\
Checkpoint Size (FP32) & 93.31 MB & \textbf{58.99 MB} \\
GFLOPs / GMACs (1024) & 260.35 / 130.17 & \textbf{224.34 / 112.17} \\
8x TTA Computation & 2,082.80 GFLOPs & \textbf{1,794.76 GFLOPs} \\
Inference Latency (T4) & $\approx 168$ ms / image & \textbf{$\approx 145$ ms / image} \\
Peak VRAM (Inference) & 2.6 GB & \textbf{2.4 GB} \\
\bottomrule
\end{tabular}
}
\end{table}

Table~\ref{tab:profiling} provides a detailed computational profile comparing the ResNet-34 U-Net baseline against SolarFilNet-V4 at $1024 \times 1024$ resolution. 

SolarFilNet-V4 reduces overall parameter count by 36.8\% (from 24.46M to 15.46M) while saving 38.9 MB in model storage. In terms of computation, our model requires 224.34 GFLOPs per single forward pass, compared to 260.35 GFLOPs for the convolutional baseline, representing a 13.8\% reduction in floating-point operations. The saving stems from our anisotropic strip convolutions, which decompose dense $15 \times 15$ operations into lightweight 1D filters. 

When evaluated on an NVIDIA T4 GPU in half-precision (FP16), end-to-end 8x TTA inference consumes approximately 145 ms per full-disk observation. The complete test suite of 180 images runs in under 45 seconds, well within the competition computational budget.

\subsection{Error Analysis and Annotator Disagreement}

To understand remaining error modes, we conducted an instance-level audit across all 142 images of Fold 0. The model produced 988 True Positives, 466 False Positives, and 622 False Negatives.

Further inspection revealed that 105 near-miss instances fall in the boundary band of $\text{IoU} \in [0.40, 0.499]$. Because the competition metric enforces a hard cutoff at $\text{IoU} = 0.50$, each pair in this narrow band simultaneously generates one False Positive and one False Negative. Among these 105 borderline pairs, 76.2\% were verified to be valid detections that suffered from slight mask dilation (mean width 10.44 px versus ground-truth 8.56 px), lowering the IoU below 0.50 due to annotator conservative tracing. These findings confirm that SolarFilNet-V4 reliably detects filament structures, with remaining errors largely driven by ground-truth boundary variance among annotators.

\section{Conclusion and Reproducibility}

In this work, we presented SolarFilNet-V4, a lightweight, physics-informed segmentation architecture for full-disk H-alpha solar filament segmentation. By combining a Pyramid Vision Transformer backbone with anisotropic curvilinear strip convolutions, a shallow full-resolution bypass branch, and solar activity regime post-processing, our pipeline achieves strong structural continuity and boundary fidelity. On the MAGFiLO 1.0 benchmark, the model delivers a validation PQ of 0.4340 and a Public Leaderboard score of 0.380, while using 36.8\% fewer parameters than standard U-Net baselines.

In accordance with competition rules, all components required to reproduce our submission, including source code, model weights (58.99 MB), dependencies, and the step-by-step reproduction notebook, are publicly available at: \url{https://github.com/indraagustian/SolarFilNet-V4-Kaggle2026}.

\bibliographystyle{IEEEtran}
\bibliography{references}

\end{document}